\documentclass[10pt,twocolumn,letterpaper]{article}

\usepackage[pagenumbers]{wacv} 

\definecolor{TableBlue}{rgb}{0.17,0.49,0.75}
\definecolor{TableRed}{rgb}{0.75,0.29,0.17}
\newcommand{\suggest}[1]{{\color{Black}#1}}
\usepackage[table]{xcolor}
\usepackage[export]{adjustbox}

\usepackage{glossaries}
\newacronym{av}{AV}{Autonomous Vehicles}
\newacronym{av2}{AV2}{Argoverse2}
\newacronym{dnn}{DNN}{Deep Neural Network}
\newacronym{lidar}{LiDAR}{Light Detection and Ranging}

\definecolor{wacvblue}{rgb}{0.21,0.49,0.74}
\usepackage[pagebackref,breaklinks,colorlinks,allcolors=wacvblue]{hyperref}
\usepackage{orcidlink}
\usepackage[capitalize]{cleveref}
\usepackage{bm}
\usepackage{makecell}
\usepackage{lipsum}
\usepackage{pifont}
\usepackage{multirow}
\usepackage{siunitx}
\usepackage[normalem]{ulem}

\def\wacvPaperID{713} 
\def\confName{WACV}
\def\confYear{2026}

\title{BlendCLIP: Bridging Synthetic and Real Domains for Zero-Shot 3D Object Classification with Multimodal Pretraining}

\author{
Ajinkya Khoche\textsuperscript{\textdagger,1,2}\orcidlink{0009-0009-6935-6797}, 
Gergő László Nagy\textsuperscript{\textdagger,1}\orcidlink{0009-0002-9479-6240}, 
Maciej Wozniak\textsuperscript{1}\orcidlink{0000-0002-3432-6151},\\
Thomas Gustafsson\textsuperscript{2}, 
Patric Jensfelt\textsuperscript{1}\orcidlink{0000-0002-1170-7162},\\
\textsuperscript{1}KTH Royal Institute of Technology \quad
\textsuperscript{2}Scania CV AB\\
{\tt\small \{khoche, glnagy, maciejw, patric\}@kth.se, \{ajinkya.khoche, thomas.gustafsson\}@scania.com}
}
\begin{document}
\maketitle

\def\thefootnote{\textdagger}\footnotetext{These authors contributed equally to this work.}\def\thefootnote{\arabic{footnote}}

\begin{abstract}
    
Zero-shot 3D object classification is crucial for real-world applications like autonomous driving, however it is often hindered by a significant domain gap between the synthetic data used for training and the sparse, noisy LiDAR scans encountered in the real-world. Current methods trained solely on synthetic data fail to generalize to outdoor scenes, while those trained only on real data lack the semantic diversity to recognize rare or unseen objects.

We introduce \textbf{BlendCLIP}, a multimodal pretraining framework that bridges this synthetic-to-real gap by strategically combining the strengths of both domains. We first propose a pipeline to generate a large-scale dataset of object-level triplets—consisting of a point cloud, image, and text description—mined directly from real-world driving data and human annotated 3D boxes. Our core contribution is a curriculum-based \textbf{data mixing} strategy that first grounds the model in the semantically rich synthetic CAD data before progressively adapting it to the specific characteristics of real-world scans. 

Our experiments show that our approach is highly label-efficient: introducing as few as 1.5\% real-world samples \suggest{per batch} into training boosts zero-shot accuracy on the nuScenes benchmark by \textbf{27\%}. Consequently, our final model achieves state-of-the-art performance on challenging outdoor datasets like nuScenes and TruckScenes, \suggest{improving over the best prior method by \textbf{19.3\%} on nuScenes}, while maintaining strong generalization on diverse synthetic benchmarks. Our findings demonstrate that effective domain adaptation, not full-scale real-world annotation, is the key to unlocking robust open-vocabulary 3D perception. Our code and dataset will be released upon acceptance on \href{https://github.com/kesu1/BlendCLIP}{https://github.com/kesu1/BlendCLIP}.


\end{abstract}

\section{Introduction}
\label{sec:introduction}

\begin{figure}[h!]
    \centering
    \includegraphics[width=0.99\columnwidth]{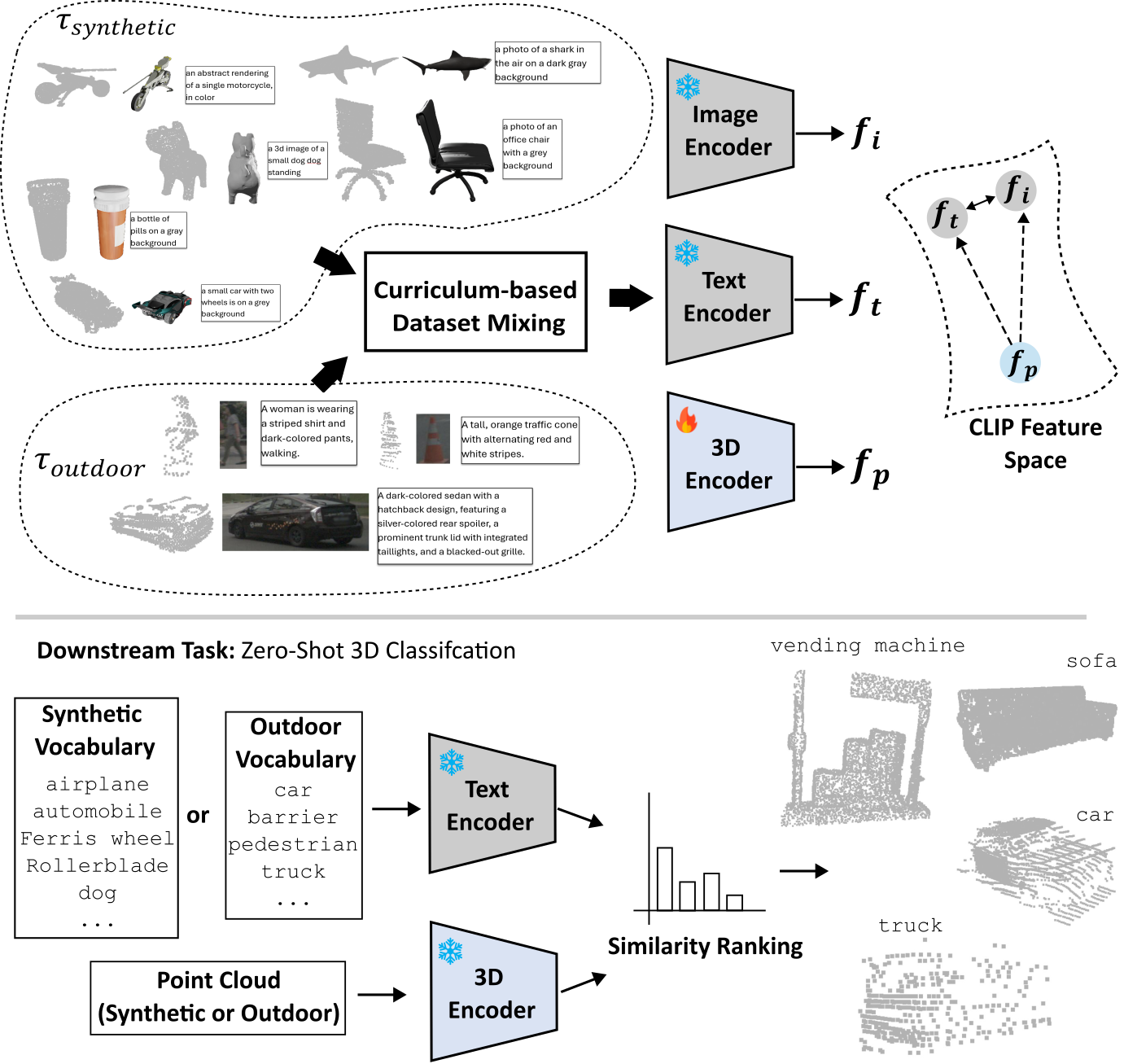}
    \caption{Overview of BlendCLIP. We construct a large-scale set of outdoor LiDAR–image–text triplets from real-world driving data and introduce a curriculum-based data mixing strategy to integrate them with existing synthetic triplets. A 3D encoder is trained to align with frozen CLIP text and image embeddings, enabling robust zero-shot 3D classification across both synthetic and real-world domains.}
    \label{fig:cover_fig}
\end{figure}

Outdoor 3D perception systems operate in complex, dynamic environments where the types of objects encountered are both diverse and unpredictable. Applications in autonomous driving and mobile robotics require models that can robustly recognize objects despite challenging conditions like sensor noise, occlusions, and the inherent sparsity of LiDAR scans. However, most existing 3D recognition pipelines rely on closed-set classification, where object categories are fixed in advance with full supervision for each class~\cite{qi2017pointnet, qi2017pointnet++, qian2022pointnext, Yu2022PointBERT}. This assumption breaks in real-world deployments, where new categories frequently emerge, label coverage is incomplete, and retraining models for every change is infeasible.

This reality motivates the critical need for \textit{zero-shot 3D object classification}—the ability to assign meaningful semantic labels to 3D objects from unseen categories without retraining. In the 2D domain, this capability has been revolutionized by large-scale vision-language models (VLMs) like CLIP~\cite{OpenAI2021CLIP}, which align images and text in a shared embedding space. However, extending this paradigm to 3D point clouds remains a significant challenge.

Early efforts have largely followed two paths, each with its own limitations. One line of work leverages large-scale synthetic CAD models to train 3D encoders aligned with VLM embeddings~\cite{Xue2024ULIP2, liu2023openshape}. While these approaches benefit from clean, dense geometries and diverse semantic labels, they struggle to bridge the sim-to-real gap, where real data (coming from LiDAR) is sparse, incomplete, and suffers from a substantial domain gap. 
The other type of methods train exclusively on real-world outdoor data, often using automatically generated labels~\cite{hess2024lidarclip, zeng2023clip2}. While tailored to the target domain, these models are constrained by the limited visual and semantic diversity of existing driving datasets, causing them to struggle with rare, long-tail object categories and perform poorly on broad zero-shot benchmarks.

In this work, we argue that robust open-vocabulary 3D classification hinges on bridging this synthetic-to-real domain gap. We propose a new framework that synergistically combines the semantic richness of synthetic data with the real-world LiDAR data. To achieve this, we first construct a large-scale dataset of LiDAR–image–text triplets from real autonomous driving data (\Cref{subsec:outdoor_triplets}). We then introduce a novel curriculum-based data mixing strategy that gradually introduces real-world LiDAR crops during training, allowing the model to adapt without forgetting the general knowledge learned from synthetic data(\Cref{subsec:data_mixing}). 

We evaluate our method on three distinct datasets spanning synthetic, urban, and highway domains. Our experiments demonstrate state-of-the-art zero-shot classification performance, achieving up to a \textbf{19.3\%} improvement on real-world data when our curriculum strategy is applied, while maintaining competitive performance on synthetic data, demonstrating strong generalization across both domains. Notably, our model's performance improves significantly with even a small fraction of real LiDAR samples~(\Cref{fig:data_efficiency}), suggesting that our approach can unlock strong generalization without requiring re-annotation of large-scale datasets. Our contributions are summarized as follows:

\begin{itemize}
    \item We introduce a pipeline for generating real-world object-level triplets (LiDAR, image, text) from large-scale autonomous driving data, and use it to construct a comprehensive triplet dataset.
    \item We introduce a curriculum-based data mixing strategy that combines a semantically rich CAD object dataset with real LiDAR crops, enabling robust generalization across synthetic and real-world domains.
    \item We evaluate our method across diverse scenarios and show significant gains on two challenging outdoor datasets (nuScenes and TruckScenes), achieving state-of-the-art zero-shot 3D classification performance.
\end{itemize}

\section{Related Work}
\label{sec:related_work}

\subsection{Multimodal Representation Learning} \label{subsec:multimodal_rep_learning}
Recent works aim to imbue 3D point-cloud encoders with the semantic richness of large-scale vision–language models, learning unified embeddings that support zero-shot classification and cross-modal retrieval of shapes. 
PointCLIP~\cite{zhang2022pointclip} projects point clouds into multi-view depth maps, encoding them with the off-the-shelf visual backbone of CLIP~\cite{OpenAI2021CLIP}. 
CLIP2Point~\cite{huang2023clip2point} renders paired depth and RGB views from infinitely dense CAD objects for contrastive pre-training, directly mapping global CLIP representations to point clouds.
ULIP~\cite{Xue2023ULIP} aligns 3D point-cloud encodings with frozen CLIP image and text embeddings, employing multimodal contrastive learning to enhance 3D understanding. Using large-scale vision-language priors through 3D-2D-text triplets, it significantly improves standard and zero-shot 3D classification. 
Follow-up works extend this by scaling up across multiple datasets, automating text caption generation~\cite{Xue2024ULIP2} and introducing strategies to filter and enrich noisy user texts~\cite{liu2023openshape}. 

\subsection{Representation Learning for Outdoor Scenes}
\label{subsec:rep_outdoor_scenes}
A primary challenge in outdoor perception is learning semantically rich representations that can be effectively applied to downstream tasks. One line of research has focused on leveraging powerful, pre-trained 2D Vision-Language Models (VLMs). These methods generate rich map representations by directly lifting 2D VLM features to 3D, enabling queriable classification, segmentation, and visual navigation~\cite{jatavallabhula2023conceptfusion, busch2024onemap, guo2024semanticgaussians}. While effective, this direct mapping process can be computationally costly, limiting scalability. To address this, other works propose large-scale pseudo-labeling; for instance, SAL~\cite{ovsep2024SAL} presents a scalable method for generating per-point instance and CLIP pseudo-labels from 2D VLMs and distills this knowledge into a 3D model. SAL-4D~\cite{zhang2025sal4d} extends this by distilling video-based pseudo-labels into temporally consistent 4D annotations. Concurrently, methods like LidarCLIP~\cite{hess2024lidarclip} and CLIP$^2$~\cite{zeng2023clip2} pioneer the direct alignment of sensor data with language embeddings, mapping raw LiDAR returns or 3D-2D pairs into the CLIP feature space to enable open-vocabulary reasoning. However, relying on existing VLMs or curated autonomous driving data can constrain generalization to novel object types not seen during training.

\suggest{\subsection{Bridging Synthetic-Real Domain Gaps}
\label{subsec:domain_generalization}

Learning transferable representations across domains remains another central challenge in 3D perception. Substantial effort has been dedicated in developing realistic simulation environments~\cite{dosovitskiy2017carla, manivasagam2020lidarsim,gaidon2016virtual}, which enable controllable variation in lighting, weather, and sensor properties, and have been widely adopted in sim-to-real transfer for detection and segmentation~\cite{zhao2021epointda, jaritz2020xmuda, xiao2022transfer,wozniak2024uada3d}. While such simulators excel at reproducing domain-specific artifacts, their semantic label spaces are typically constrained to road users and traffic infrastructure. In parallel, internet-scale CAD datasets~\cite{chang2015shapenet,Deitke2023Objaverse} spanning thousands of categories have emerged, providing semantic diversity essential for open-vocabulary classification. 
Consistent object-level correspondences afforded by CAD assets further make object-centric synthetic data indispensable for multimodal contrastive pretraining and for learning transferable, language-grounded 3D representations.

However, a key difficulty remains when combining synthetic and real data indiscriminately:
models may overfit to the \textit{domain} as the simplest separating signal rather than learning semantically meaningful features~\cite{torralba2011unbiased}. Prior works~\cite{Bengio2009curriculum, wang2021survey} have investigated training from easy to hard via a pacing function, or introducing intermediate stages~\cite{hsu2020progressive} to bridge large domain gaps. Such strategies have proven effective in object detection and semantic segmentation, where staged exposure to target pseudo-labels avoids the failure modes of naive one-shot mixing~\cite{banitalebi2023ebcdet, soviany2022curriculum, wang2024curriculum}. 

We show that \textit{curriculum learning} offers a principled solution to prevent the model from collapsing onto dataset cues, keeping high performance over all training datasets.

}

\section{Method}\label{sec:method}

\begin{figure*}[htbp] 
    \centering
    \includegraphics[width=1.95\columnwidth]{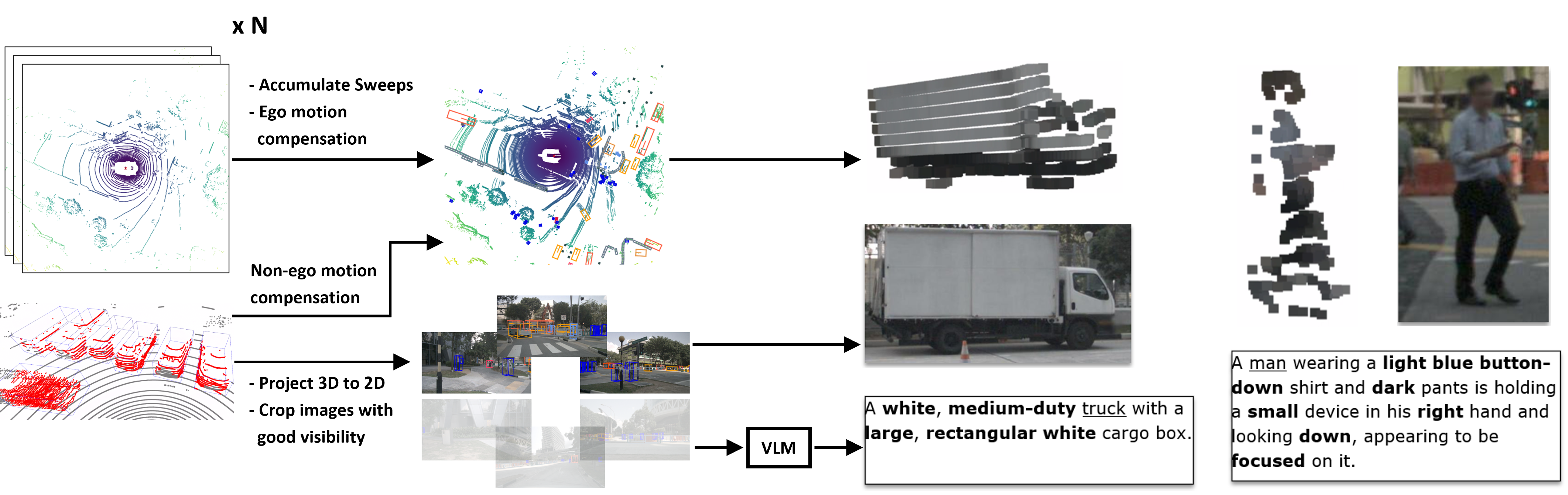} 
    \caption{Overview of outdoor triplet generation. Given time-synchronized sensor data, we apply multi-sweep fusion and motion compensation to extract dense object-level point clouds from annotated 3D boxes. Image crops are obtained by projecting the box into synchronized camera views and selecting those with high visibility. Each image is then captioned using a vision-language model to form point cloud–image–text triplets used in training. \underline{Underlined} denotes the subject, while \textbf{bold} indicates fine-grained descriptions.}
    \label{fig:triplet_schematic}
\end{figure*}


We build upon the pre-training strategy of ULIP-2~\cite{Xue2024ULIP2}, which learns a 3D encoder by aligning its output with frozen image and text CLIP embeddings using 3D-image-text triplets (\Cref{subsec:background}). The resulting encoder can be used for various open-vocabulary downstream tasks. We extend this framework with two main contributions. 
First, we construct a \textit{novel outdoor triplet dataset} using synchronized sensors and human-annotated bounding boxes (\Cref{subsec:outdoor_triplets}). 
Second, we introduce a \textit{data-mixing strategy for joint training} on synthetic and real-world triplets (\Cref{subsec:data_mixing}). 

\subsection{Background: ULIP-2 Pretraining} \label{subsec:background}
ULIP-2~\cite{Xue2024ULIP2} introduces a tri-modal pretraining strategy for learning open-vocabulary 3D representations. Given a 3D object, the pipeline samples a point cloud, renders a view, and generates a caption using a vision-language model (BLIP-2). The 3D encoder is trained to align its output with the embeddings of the rendered image and caption using a contrastive objective.
Formally, let $f_P$, $f_I$, and $f_T$ denote the encoders for point cloud, image, and text respectively. The total loss combines two contrastive alignment terms:
\begin{equation}
\mathcal{L} = \mathcal{L}_{P2I} + \mathcal{L}_{P2T}
\end{equation}
Each $\mathcal{L}_{P2X}$ is a symmetric InfoNCE~\cite{oord2019infonce} loss computed over a batch, encouraging the 3D embedding to match both the image and text embeddings of the same object, while pushing apart mismatched pairs. The image and text encoders are initialized from CLIP and kept frozen. This enables ULIP-2 to leverage vision-language priors without requiring 3D annotations.

\subsection{Outdoor Triplet Generation} \label{subsec:outdoor_triplets}
To address the inability of ULIP‑2’s synthetic-data-driven triplet sampling to generalize to real‐world LiDAR (shown in \Cref{tab:zero_shot_classification_detailed})—where sensor noise, non-uniform point densities, and annotation inconsistencies are pervasive, we propose a training data generation pipeline tailored for autonomous driving scenarios. This section details how we construct high-quality triplets from time-synchronized multimodal sensor data in order to combat those issues.

Let $\mathcal{D}=\{(\mathcal{P}_t, \mathcal{I}^k_t, \mathcal{B}_t)\}_{t=1}^T$ denote the set of time-synchronized multimodal sensor data collected over $T$ timesteps, where $\mathcal{P}_t$ is the raw LiDAR point cloud captured by one or more sensors, $\mathcal{I}^k_t$ is the set of synchronized RGB images from $K$ calibrated cameras, and $\mathcal{B}_t$ is the set of annotated 3D bounding boxes at time $t$. 
To construct high-quality object-level point cloud at a reference time $t_0$, we first aggregate LiDAR measurements from the current frame and the past $N$ sweeps by compensating for ego-vehicle motion. Each LiDAR point cloud $\mathcal{P}_t$ (represented in homogeneous coordinates) is transformed into the sensor frame at time $t_0$ using a chain of rigid-body transformations:
\begin{equation}
{}^{s_{t_0}}\mathcal{P}_t = T^{s_{t_0}}_{e_{t_0}} \, T^{e_{t_0}}_{g} \, T^{g}_{e_t} \, T^{e_t}_{s_t} \, \mathcal{P}_t,
\end{equation}
where $T^{a}_{b} \in \mathrm{SE}(3)$ is the transformation from frame $b$ to frame $a$, $s_t$ and $e_t$ denote the sensor and ego-vehicle frames at time $t$ and $g$ is the global coordinate frame.
To further account for objects moving independently to ego-motion, we estimate per-object linear and rotational motion and compensate each sweep before merging. 

Let $M$ denote the interval between annotated frames, i.e., $\mathcal{B}_{t}$ is available only at timestamps $t \in \{t_0, t_0 - M, t_0 - 2M, \dots\}$. To compensate for object motion during sweep accumulation, we interpolate object pose (center and orientation) at each intermediate timestamp $t \in [t_0 - M, t_0]$ using bounding boxes from $\mathcal{B}_{t_0 - M}$ and $\mathcal{B}_{t_0}$.
For an object $i$ with annotated centers $\bm{c}_i(t_0 - M), \bm{c}_i(t_0) \in \mathbb{R}^3$ and orientations $\bm{q}_i(t_0 - M), \bm{q}_i(t_0) \in \mathrm{SO}(3)$, we define the linearly interpolated center at time $t$ as:
\begin{equation}
\bm{c}_i(t) = \bm{c}_i(t_0) + \left( \frac{t - t_0}{M} \right) \cdot (\bm{c}_i(t_0) - \bm{c}_i(t_0 - M)),
\end{equation}
and the interpolated orientation via Spherical Linear Interpolation (\texttt{slerp}) as:
\begin{equation}
\bm{q}_i(t) = \mathrm{slerp}\left(\bm{q}_i(t_0), \bm{q}_i(t_0 - M), \frac{t_0 - t}{M} \right),
\end{equation}
where
\begin{equation}
\mathrm{slerp}(\bm{q}_1, \bm{q}_2, \alpha) = \frac{\sin[(1 - \alpha)\theta]}{\sin \theta} \bm{q}_1 + \frac{\sin(\alpha \theta)}{\sin \theta} \bm{q}_2
\end{equation}
with $\quad \theta = \cos^{-1}(\bm{q}_1 \cdot \bm{q}_2)$. Thereafter we estimate the object's constant linear velocity over the interval $[t_0 - M, t_0]$ as:
\begin{equation}
\bm{v}_i = \frac{\bm{c}_i(t_0) - \bm{c}_i(t_0 - M)}{M}.
\end{equation}
At sweep time $t$, we crop points for object $i$ using the interpolated box and forward-warp them to the reference frame $t_0$ using the velocity $\bm{v}_i$ and inverse rotation:
\begin{equation}
\bm{p}_i^{(t_0)} = \bm{R}_i(t_0)^\top \left[ \bm{p}_i^{(t)} + (t_0 - t) \cdot \bm{v}_i - \bm{c}_i(t_0) \right],
\end{equation}
where $\bm{R}_i(t_0) \in \mathrm{SO}(3)$ is the rotation matrix corresponding to $\bm{q}_i(t_0)$. This transformation brings all object points into a canonical pose at $t_0$ for dense fusion across sweeps.

To obtain corresponding RGB views, each object’s 3D bounding box $\bm{b}_i(t_0)$ is projected into all $K$ camera frames available at time $t_0$. Using standard pinhole projection with available extrinsic and intrinsic calibration, the 3D box corners are mapped to 2D and a tight axis-aligned bounding box is fitted to define the image crop. To ensure crop quality, \suggest{it is done only if (i) all six corners of $\bm{b}_i^{(t_0)}$ project inside the image plane, and (ii) the object is above a visibility threshold. This yields the set of valid crops:
\begin{equation}
    \mathcal{I}_i = \bigcup_{t=t_{\text{start}}}^{t_{\text{end}}} 
    \Bigl\{
    \text{Crop}_{\text{AABB}}\!\left(\bm{I}_t^{(j)}, \bm{b}_i(t)\right) \; \big| \; \forall j \in \mathcal{V}_i^{(t)} 
    \Bigr\},
\end{equation}
where $\mathcal{V}_i^{(t)} \subseteq  \{1, \dots, K\} $ indexes the views that satisfy the visibility and projection constraints at annotated time $t$, and $\mathrm{Crop}_{\mathrm{AABB}}$ denotes the projection-then-cropping operation. $t_{start} \leq t_{end}$ denotes the first and last time stamps of the occurrence of the instance $i$.}

%

To obtain natural language descriptions for each object, we use a pretrained image–text model to generate \suggest{captions for all the image crops in $\mathcal{I}_i$, yielding the set of captions:
\begin{equation}
    \mathcal{Y}_i = \Bigl\{\bm{y}_i^{(j)}=\text{VLM}\!\left(\bm{i}_i^{(j)}\right) \; \big| \; \forall \bm{i}_i^{(j)} \in \mathcal{I}_i \Bigr\},
\end{equation}
where $\text{VLM}$ denotes the captioning with the image–text model operation.}
Combining the fused object point cloud, available image crops, and their generated caption, we define the set of available triplets for object $i$ at time $t_0$ as
\suggest{
\begin{equation}
    \mathcal{T}_i^{(t_0)}=
    \Bigl\{
    \left( \bm{p}_i^{(t_0)}, \bm{i}_i^{(j)}, \bm{y}_i^{(j)} \right) \; \big| \; \bm{i}_i^{(j)} \in \mathcal{I}_i, \; \bm{y}_i^{(j)} \in \mathcal{Y}_i 
    \Bigr\},
\end{equation}
which forms multiple training instances for multimodal alignment, allowing the model to observe multiple viewing angles of the same object.}
\subsection{Curriculum for Bridging the Domain Gap}
\label{subsec:data_mixing}

\suggest{A central challenge in mixing synthetic CAD and outdoor LiDAR data is that their distributions are trivially distinguishable: models can easily classify dataset identity~\cite{torralba2011unbiased}, and often exploit such shortcuts instead of learning semantic invariances~\cite{geirhos2020shortcut}. In contrastive learning this issue is amplified, since the InfoNCE objective is sensitive to the negative sample distribution~\cite{saunshi2019theoretical}; if most negatives come from another domain, the model may push apart semantically similar samples simply because they differ in dataset identity. Naively combining synthetic and real triplets from the start therefore risks encouraging spurious domain separation, especially early in training when the learning rate is highest and representations consolidate quickly.

To mitigate this, we adopt a \textit{curriculum learning strategy} that gradually increases the contribution of outdoor triplets. The intuition is that the model should first establish a broad semantic foundation from the diverse, clean CAD data, and then progressively adapt to domain-specific artifacts as training stabilizes. This schedule avoids catastrophic forgetting of semantic diversity while improving transfer to real-world LiDAR, as confirmed by our ablations (\Cref{subsec:ablation_studies}).} This is implemented as a two-phase schedule:

\begin{enumerate}
    \item \textbf{Semantic Priming Phase:} For an initial number of warm-up epochs ($W_e$), the model is trained exclusively on synthetic CAD-based triplets. This enables the 3D encoder to learn a rich, general-purpose feature space that spans a wide variety of object classes, without the corrupting influence of sensor noise or sparsity.

    \item \textbf{Domain Adaptation Phase:} Following the warm-up, we progressively introduce the real-world LiDAR triplets into the training batches. The fraction of real-world samples, $r(e)$, is gradually increased from zero to a target maximum, $r_{\max}$, over the remaining epochs. This gradual shift allows the model to adapt its learned semantic features to the geometric and statistical properties of real LiDAR data in a stable manner.
\end{enumerate}
\suggest{We employ a linear mixing function, as it represents the simplest approach to data mixing.}
The mixing ratio at epoch $e$ is formally defined as:
\begin{equation}
r(e) =
\begin{cases}
0, & \text{if } e < W_e, \\
\frac{e - W_e}{T_e - W_e} \cdot r_{\max}, & \text{if } W_e \le e \le T_e.
\end{cases}
\end{equation}
This delicate balance is critical; as our ablations in \Cref{tab:data_mixing_ratio} demonstrate, too few real samples fail to bridge the domain gap, while too many can cause the model to "forget" the semantic diversity learned from the synthetic data. By sampling from both domains within the same batch, the contrastive loss benefits from both diverse, "easy" negatives (from CAD) and domain-specific, "hard" negatives (from LiDAR), promoting a more robust final embedding space.

To maintain a consistent training schedule despite the evolving sampling ratios, we precompute the number of iterations per epoch. Let $N_{\text{CAD}}$ be the number of unique CAD samples, $B$ the batch size per GPU, and $N_{\text{GPU}}$ the number of GPUs. To achieve a target expected coverage $\psi$ over the CAD dataset (i.e., the probability of having seen each sample at least once), we draw a total of
\begin{equation}
N_{\text{iter}} = \left\lceil \frac{ N_{\text{CAD}} \cdot \ln\left( \frac{1}{1 - \psi} \right) }{ N_{\text{GPU}} \cdot B } \right\rceil
\end{equation}
batches per epoch, assuming sampling with replacement. This setup, derived from the Coupon Collector’s Problem~\cite{boneh1997coupon}, ensures fair exposure to the synthetic dataset throughout training.

\section{Experimental Setup}\label{sec:expt_setup}

\subsection{Datasets} \label{subsec:datasets}

\paragraph{CAD-Based Pretraining Data:} We use ULIP-Objaverse~\cite{Xue2024ULIP2}, a large-scale dataset of 3D–image–text triplets derived from Objaverse 1.0~\cite{Deitke2023Objaverse}, for pretraining. Each sample consists of a point cloud rendered from a 3D mesh, an RGB image rendered from one of 12 fixed viewpoints, and a caption generated by BLIP-2~\cite{li2023blip2}. This synthetic dataset contains $\sim$781k triplets and provides broad semantic coverage across thousands of object categories. We also use Objaverse-LVIS, a curated subset aligned to the LVIS taxonomy~\cite{Gupta2019LVIS}, spanning across 1156 categories, for zero-shot evaluation.

\paragraph{Outdoor Triplet Construction:} We extract real-world 3D–image–text triplets from the nuScenes~\cite{Caesar2020nuscenes} and TruckScenes~\cite{fent2024man} datasets, following the procedure described in~\Cref{subsec:outdoor_triplets}. The resulting \textit{nuScenes-triplets} dataset (or nuS for brevity) is used for both training and evaluation. 
\suggest{For training, we aggregate 10-sweeps to densify each LiDAR crop and filter out instances with fewer than 150 points. For evaluation, we follow the official nuScenes object detection benchmark rules, removing boxes that exceed the class-specific detection range. Finally,}
we map the original 23 classes to 10 categories, and filter out sparse instances, following the official nuScenes benchmark. 
\textit{TruckScenes-triplets} (or TruckS) is used for zero-shot evaluation only and includes the same categories as nuScenes-triplets, along with an additional class: \texttt{traffic sign}. It poses a greater challenge due to annotations extending up to 150m, where objects are especially sparse and occluded. Unlike nuScenes, its dense point clouds are obtained directly via multi-sensor fusion across six 64-beam LiDARs, without multi-sweep accumulation.

\paragraph{Dataset Scale:} The combined training and evaluation sets span over one million diverse 3D objects, ensuring sufficient coverage for both pretraining and zero-shot generalization. Dataset statistics are provided in~\Cref{tab:dataset_sample_size}.
\begin{table}[htbp]
    \centering
    \renewcommand{\arraystretch}{1.2}
    \small
    \resizebox{0.7\columnwidth}{!}{%
    \begin{tabular}{lcc}
    \toprule
    Dataset & Training & Validation \\
    \midrule
    Objaverse~\cite{Xue2024ULIP2,Deitke2023Objaverse} & $\sim$781k & $\sim$46k \\
    nuScenes-triplets & $\sim$220k & $\sim$44k \\
    TruckScenes-triplets & -- & $\sim$30k \\
    \bottomrule
    \end{tabular}
    }
    \caption{Sample sizes of datasets used for training and evaluation.}
    \label{tab:dataset_sample_size}
\end{table}

\begin{table*}[t]
    \centering
    \renewcommand{\arraystretch}{1.4}
    \small
    \resizebox{1.8\columnwidth}{!}{%
    \begin{tabular}{l c ccc c ccc}
        \toprule
        \multirow{2}{*}{Method} & \multirow{2}{*}{Pre-train Dataset} & \multicolumn{3}{c}{Class-wise (\%) ↑} & & \multicolumn{3}{c}{Object-wise (\%) ↑} \\
        \cmidrule(lr){3-5} \cmidrule(lr){7-9}
        & & Objaverse-LVIS~\cite{Deitke2023Objaverse} & nuS & TruckS & & Objaverse-LVIS~\cite{Deitke2023Objaverse} & nuS & TruckS \\
        \midrule
        PointCLIP$^*$~\cite{zhang2022pointclip} & nuS & -- / -- & 21.1 / -- & -- / -- & & -- / -- & -- / -- & -- / -- \\
        CLIP2Point$^*$~\cite{huang2023clip2point} & nuS & -- / -- & 24.0 / -- & -- / -- & & -- / -- & -- / -- & -- / -- \\
        CLIP$^{2*}$~\cite{zeng2023clip2} & nuS & -- / -- & 28.7 / -- & -- / -- & & -- / -- & -- / -- & -- / -- \\
        LidarCLIP$^{\dagger}$~\cite{hess2024lidarclip} & nuS & -- / -- & 10.1 / 50.1 & -- / -- & & -- / -- & 12.5 / 59.0 & -- / -- \\
        ULIP-2$^{\dagger}$~\cite{Xue2024ULIP2} & nuS & 0.2 / 0.8 & \underline{43.8} / \textbf{81.1} & \underline{21.0} / \underline{62.7} & & 0.2 / 0.9 & \underline{46.6} / \underline{87.7} & 17.9 / \underline{67.3} \\
        ULIP-2$^{\dagger}$~\cite{Xue2024ULIP2} & Objaverse~\cite{Deitke2023Objaverse} & \textbf{35.7} / \textbf{64.2} & 9.9 / 56.6 & 8.5 / 48.1 & & \textbf{47.8} / \textbf{75.8} & 11.6 / 53.8 & 4.1 / 23.4 \\
        \midrule
        BlendCLIP (Ours) & Objaverse~\cite{Deitke2023Objaverse}+nuS & \underline{34.5} / \underline{63.6} & \textbf{48.0} / \underline{79.5} & \textbf{25.5} / \textbf{63.5} & & \underline{46.7} / \underline{75.2} & \textbf{59.7} / \textbf{87.7} & \textbf{43.5} / \textbf{85.5} \\
        \bottomrule
    \end{tabular}
    }
    \caption{Zero-shot classification results on synthetic and outdoor benchmarks. Each cell reports Top-1 / Top-5 accuracy (\%). $*$: reported by authors of CLIP$^2$~\cite{zeng2023clip2}. $^{\dagger}$: reproduced by us. Best results in bold, second-best are underlined.}
    \label{tab:zero_shot_classification_detailed}
\end{table*}
\vspace{-1em}

\subsection{Evaluation Protocol} \label{subsec:evaluation_protocol} 

\suggest{We evaluate the learned 3D encoder through zero-shot object classification. Following prior work~\cite{zeng2023clip2,Deitke2023Objaverse,hess2024lidarclip}, we report object-wise (instance-level) Top-1 accuracy, and additionally include class-wise (mean per-class) accuracy as well as Top-5 variants of both metrics.}
Following CLIP-style evaluation~\cite{OpenAI2021CLIP}, we embed both 3D point clouds and class name prompts into a shared feature space and compute cosine similarity between them for classification.

For evaluation on Objaverse-LVIS, we follow ULIP~\cite{Xue2023ULIP} and apply a diverse set of 64 natural language prompts per class, including 63 CLIP-style templates and one 3D-specific prompt: \texttt{"a point cloud model of \{\}."}. Each prompt is embedded using the text encoder, and the final class representation is computed by averaging the resulting embeddings.
For real-world datasets (nuScenes and TruckScenes), we adopt the minimalist strategy of CLIP$^2$~\cite{zeng2023clip2}, inserting each class name into a single fixed prompt: \texttt{"point cloud of \{\}"}. 

\subsection{Implementation Details} \label{subsec:implementation_details}

\suggest{We base our implementation on ULIP-2~\cite{Xue2024ULIP2}}. We adopt Point-BERT~\cite{Yu2022PointBERT} as our default 3D encoder and CLIP-ViT/B-16~\cite{OpenAI2021CLIP} as the default frozen image and text encoders. Only the 3D backbone is trained.
During training, the warm up epoch $W_e$ and total epochs $T_e$ are chosen to be 1 and 250 respectively. The default mixing ratio $r_{max}$ is chosen to be 30\%. To ensure consistent schedule during dataset mixing, the expected coverage $\psi$ is set to 0.8. The training is performed on 8×A100 GPUs
For additional implementation details, please see~\Cref{supp:additional_implementation_details}.

\section{Experimental Results}\label{sec:expt_results}

\subsection{Quantitaive Results} \label{subsec:quantitative_results}

A key goal of this work is to achieve strong open-vocabulary generalization across both class diversity and domain variation. To this end, we evaluate our model on three complementary datasets, each of which tests a distinct axis of generalization.

\textit{Objaverse-LVIS} evaluates the model’s ability to recognize a large vocabulary of fine-grained object categories under clean synthetic conditions. With over 1000 classes and long-tailed distribution aligned to the LVIS taxonomy, success on this dataset indicates effective learning of semantic priors and intra-class diversity. However, strong performance here alone is insufficient to claim zero-shot robustness in real-world settings.

\textit{nuScenes-triplets} introduces real-world LiDAR characteristics, including occlusion, sparsity, and range-dependent noise. It contains only a small number of annotated classes common in driving environments. Performance on nuScenes thus reflects the model’s ability to transfer from clean CAD training data to noisy, cluttered outdoor scenes, a key requirement for real-world deployment. \textit{TruckScenes-triplets} further pushes domain shift by evaluating on different hardware (six 64-beam LiDARs), sensor placement, and scene context (primarily highways), \suggest{and is never observed during training}. This setting tests the model’s ability to generalize beyond the training distribution — not just to new classes or clutter, but to entirely new sensing setups and environments.

\suggest{\Cref{tab:zero_shot_classification_detailed} summarizes the results. ULIP-2 trained on synthetic data, achieves strong performance on Objaverse-LVIS, but degrades sharply on outdoor benchmarks due to the domain gap. Conversely, ULIP-2 trained on nuScenes obtains stronger accuracy in outdoor scenes (43.8/46.6 Top-1 class-/object-wise) but collapses on the diverse class set of Objaverse-LVIS (0.2\%) due to limited semantic coverage in training. BlendCLIP consistently balances these extremes: it matches Objaverse performance (34.5/46.7) while substantially improving on both outdoor datasets. On nuScenes, BlendCLIP improves over the best baseline by +4.2\% class-wise and +13.1\% object-wise Top-1 accuracy, and on TruckScenes it more than doubles accuracy compared to Objaverse-only pretraining (43.5 vs. 17.9 object-wise Top-1). These results demonstrate that our curriculum-based mixing of CAD-based synthetic and outdoor triplets successfully preserves semantic breadth while providing superior transfer to challenging real-world settings.
}
\suggest{\paragraph{Evaluation on Unseen Classes:}
To further assess zero-shot generalization, we withhold four nuScenes categories---\texttt{pedestrian}, \texttt{traffic cone}, \texttt{bus}, and \texttt{motorcycle}---from all outdoor training triplets, ensuring they are never observed during contrastive learning. These categories remain present in Objaverse, enabling evaluation of cross-domain transfer from synthetic to outdoor scenes. \Cref{tab:unseen_classes} reports top-1 accuracy for these held-out classes under two conditions: (i) \textit{All Objects}, where all valid ground-truth instances are included, and (ii) \textit{Highly Visible Objects}, where only dense crops with at least 150 LiDAR points are retained.  
%

\begin{table}[h]
    \centering
    \renewcommand{\arraystretch}{1.4}
    \small
    \resizebox{0.99\columnwidth}{!}{%
    \begin{tabular}{lcccccccc}
        \toprule
        \multirow{2}{*}{Method} &
        \multicolumn{4}{c}{All Objects (\%) ↑} &
        \multicolumn{4}{c}{Highly Visible Objects (\%) ↑} \\
        \cmidrule(lr){2-5} \cmidrule(lr){6-9}
        & Ped. & T.C. & Bus & Motor. & Ped. & T.C. & Bus & Motor. \\
        \midrule
        ULIP-2~\cite{Deitke2023Objaverse} & \textbf{43.6} & \textbf{18.4} & 10.0 & \textbf{15.6} & 6.1 & \textbf{57.3} & 10.1 & 18.7 \\
        BlendCLIP (Ours) & 18.9 & 11.4 & 26.0 & 12.7 & \textbf{73.0} & 54.9 & \textbf{29.3} & \textbf{23.4} \\
        \bottomrule
    \end{tabular}
    }
    \caption{Top-1 zero-shot classification accuracy (\%) on held-out nuScenes classes: Pedestrian (Ped.), Traffic Cone (T.C.), Bus, and Motorcycle (Motor.). Results are reported for \textit{All Objects} in the validation set and for \textit{Highly Visible Objects} (instances with more than 150 points).}
    \label{tab:unseen_classes}
\end{table}
Under the \textit{All Objects} condition, ULIP-2 achieves higher accuracy on most categories, suggesting that when objects are extremely sparse, synthetic-only pretraining provides a stronger prior than mixed-domain training. However, restricting evaluation to \textit{Highly Visible Objects} reveals a very different picture: BlendCLIP exhibits large improvements across categories, with particularly notable gains for \texttt{pedestrian} (+67\% absolute) and \texttt{bus} (+19.2\%). This trend is consistent with our training setup, where only dense crops were included to ensure reliable supervision, thereby biasing the model toward strong recognition performance when sufficient points are available.
We expect that including less dense crops during training could reduce the gap under sparse conditions, but may also introduce noisier supervision; we leave this as an avenue for future work.
In safety-critical applications such as autonomous driving, recognition accuracy in the near-field is most crucial, and it is precisely in this regime that BlendCLIP demonstrates its greatest strength.
}

\begin{figure}[h!]
    \centering
    \includegraphics[width=0.65\columnwidth]{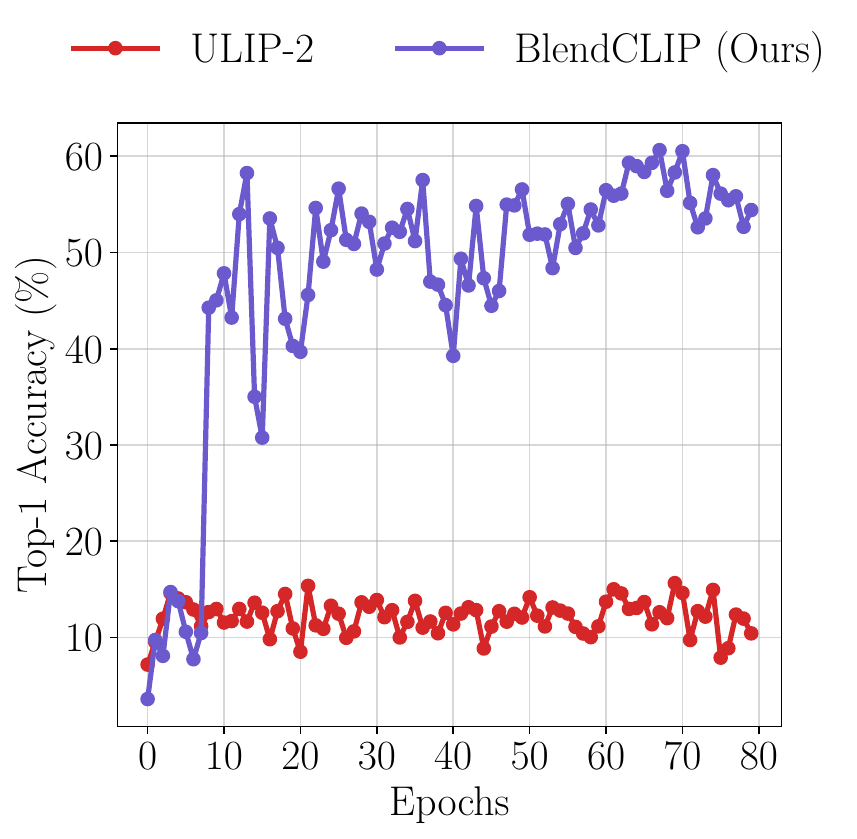}
    \caption{Comparison of top-1 accuracy on nuScenes-triplets validation set over the first \suggest{80} training epochs. Our method (BlendCLIP) exhibits a sharp performance jump at epoch 8,
    after adding just one sample per GPU,
    while the baseline trained only on synthetic data remains flat.}
    \label{fig:data_efficiency}
    \vspace{-1em}
\end{figure}

\paragraph{Label Efficiency:}
We assess how efficiently our method leverages real-world annotations by tracking \suggest{mean class-wise} top-1 classification accuracy on the nuScenes-triplets validation set across the first \suggest{80} training epochs (\Cref{fig:data_efficiency}). We compare two models: (i) ULIP-2 trained solely on synthetic CAD triplets, and (ii) our model with dataset mixing, where outdoor LiDAR triplets are gradually introduced after a warmup phase.

During the first 7 epochs, both models receive only synthetic data, resulting in similarly low accuracy (6–14\%). However, in epoch 8, our model incorporates just one sample per GPU (equivalent to 1.5\% real-world samples per batch), and performance jumps dramatically by 27\%. In contrast, the baseline remains flat throughout. This striking improvement, triggered by minimal exposure to real-world triplets, demonstrates that minimal outdoor supervision is sufficient to activate meaningful domain adaptation, provided the model has been semantically primed on diverse synthetic data. This result highlights the label efficiency of our approach, especially valuable in domains where dense manual annotations are costly or scarce.



\subsection{Ablation Studies} \label{subsec:ablation_studies}

\suggest{
\paragraph{Effectiveness of Data Mixing:} 
We analyze our design choice of curriculum-based mixing by comparing against two alternative strategies.  \textit{Static mixing} fixes the ratio of synthetic and outdoor triplets at the input stage, while \textit{two-step training} first pre-trains on Objaverse and then continues training on nuScenes from this checkpoint. Results are summarized in Table~\ref{tab:effect_data_mixing_strategies}. Static mixing emerges as a strong baseline, performing well on both Objaverse-LVIS and nuScenes. Two-step training achieves slightly higher performance on nuScenes, indicating that Objaverse pretraining provides useful initialization. However, it suffers catastrophic forgetting on Objaverse-LVIS, dropping to nearly zero accuracy. The model adapts to the ten nuScenes categories, but loses the semantic diversity necessary for open-vocabulary recognition across thousands of classes.

\begin{table}[h]
    \centering
    \renewcommand{\arraystretch}{1.2}
    \small
    \resizebox{0.8\columnwidth}{!}{%
    \begin{tabular}{lcccc}
        \toprule
        \multirow{2}{*}{Mixing Strategy} & 
        \multicolumn{2}{c}{Objaverse-LVIS~\cite{Deitke2023Objaverse} (\%) ↑} & 
        \multicolumn{2}{c}{nuS (\%) ↑} \\
        \cmidrule(lr){2-3} \cmidrule(lr){4-5}
        & Cls. & Obj. & Cls. & Obj. \\
        \midrule
        Static & 34.0 & 46.3 & 44.7 & 48.9 \\
        Two-step & 0.1 & 0.2 & 46.7 & 52.7  \\
        Curriculum (Ours) & \textbf{34.5} & \textbf{46.7} & \textbf{48.0} & \textbf{59.7} \\
        \bottomrule
    \end{tabular}}
    \caption{Effect of different mixing strategies on Top-1 zero-shot classification.}
    \label{tab:effect_data_mixing_strategies}
    \vspace{-1em}
\end{table}

Our curriculum strategy achieves the best of both worlds: it preserves Objaverse-LVIS performance (matching or slightly exceeding static mixing) while also delivering the highest accuracy on nuScenes. 
Compared to static mixing, curriculum improves nuScenes performance by +3.3\% (class-wise) and +10.8\% (object-wise). 
These results validate that gradually increasing the proportion of outdoor triplets is more effective than either static or staged alternatives, ensuring both semantic breadth and strong domain adaptation.

}

\paragraph{Effect of Data Mixing Ratio:} 
In~\Cref{tab:data_mixing_ratio}, we investigate how the maximum outdoor mixing ratio $r_{\max}$ affects zero-shot performance across domains. As expected, increasing the proportion of outdoor LiDAR crops leads to a consistent decrease in accuracy on Objaverse-LVIS. This is likely due to the reduced presence of high-resolution CAD objects in the training batches, which limits the model's exposure to long-tailed fine-grained categories.

\begin{table}[h!]
    \centering
    \renewcommand{\arraystretch}{1.4}
    \small
    \resizebox{0.75\columnwidth}{!}{%
    \begin{tabular}{lcccc}
        \toprule
        $r_{max}$ & Objaverse-LVIS~\cite{Deitke2023Objaverse} (\%) ↑ & nuS (\%) ↑ & TruckS (\%) ↑ \\
        \midrule
        10 & \textbf{47.9} & 47.7 & 24.9 \\
        20 & 47.3 & 47.8 & 23.3 \\
        30 & 46.7 & \textbf{48.0} & \textbf{25.5} \\
        40 & 45.7 & 46.4 & 23 \\
        50 & 45 & 45.2 & 22.2 \\
        \bottomrule
    \end{tabular}
    }
    \caption{Effect of dataset mixing ratio $r_{max}$ (\%) on average class-wise Top-1 zero-shot classification accuracy (\%). Higher indicates better performance.}
    \label{tab:data_mixing_ratio}
\end{table}


On nuScenes and TruckScenes triplets, accuracy increases as $r_{\max}$ grows, peaking at $30\%$ before declining. This indicates that while some real-world supervision is essential for bridging the domain gap, too much can erode the semantic generalization learned from CAD. Overall, $r_{\max} = 30\%$ provides the best trade-off, yielding the strongest outdoor performance with only a moderate drop on Objaverse-LVIS. These results validate our curriculum-style mixing strategy and confirm that only a modest proportion of real-world supervision is sufficient to drive effective domain transfer without sacrificing generalization to unseen or synthetic classes. \\

\begin{figure}[htbp]
    \centering
    \includegraphics[width=0.99\columnwidth]{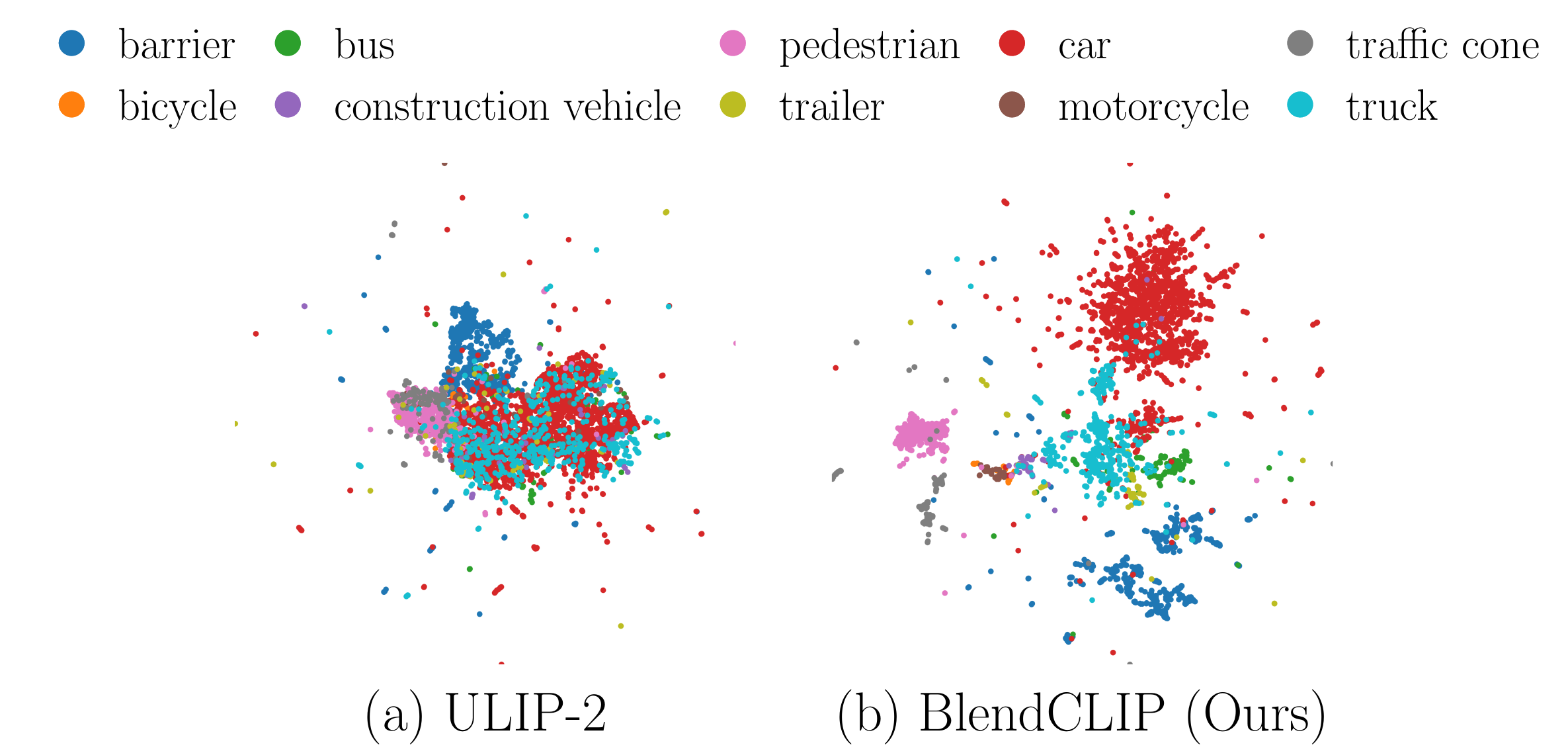}
    \caption{2D UMAP projection of CLIP-based 3D features on nuScenes-triplets validation set. (a) ULIP‑2 pretrained solely on synthetic CAD models shows poor class separation. (b) Our model, trained with data mixing, exhibits clearer clustering of real-world object categories. Colors denote ground-truth classes.}
    \label{fig:qualitative_umap}
    \vspace{-1em}
\end{figure}

\noindent \textbf{Feature Visualization:} To provide qualitative insight into how our dataset mixing strategy shapes the 3D-text embedding space, we use UMAP (Uniform Manifold Approximation and Projection)~\cite{mcinnes2018umap} to project high-dimensional point-cloud features into 2D. 
In order to ensure meaningful geometry and exclude overly sparse crops, we restrict the visualization to object instances containing more than 150 LiDAR points, in the nuScenes-triplets validation set.
As shown in~\cref{fig:qualitative_umap}, prior to dataset mixing, clusters exhibit significant overlap among classes, indicating poor separation of real-world categories. After mixing, clusters become more distinct, especially for real-world vehicle classes, showing improved alignment with semantic categories. This increased cluster clarity reflects better cross-domain discrimination, supporting our quantitative gains in zero-shot accuracy.
%

\section{Conclusion} \label{sec:conclusion}

We present BlendCLIP, a multimodal pretraining framework for zero-shot 3D object classification in autonomous driving. Our method tackles the synthetic-to-real domain gap by integrating real-world object-level triplets, composed of LiDAR, image, and text, with semantically rich synthetic data through a curriculum-based data mixing strategy. Extensive experiments show that our approach substantially improves zero-shot performance on challenging outdoor datasets like nuScenes and TruckScenes, while preserving strong generalization on synthetic benchmarks. Remarkably, the model demonstrates strong label efficiency, achieving large performance gains with modest real-world supervision.

While our dataset mixing strategy effectively bridges domain gaps, it currently relies on a fixed curriculum schedule and handcrafted mixing ratio. Future work may explore adaptive mixing strategies conditioned on training dynamics or domain confidence. Additionally, although our real-world triplet construction is scalable, it still depends on labeled 3D bounding boxes; reducing this reliance via automatic object discovery or active learning remains 
remains an open direction.
{
    \small
    \bibliographystyle{ieeenat_fullname}
    \bibliography{main}
}

\clearpage  

\setcounter{figure}{0}
\setcounter{table}{0}
\setcounter{equation}{0}
\setcounter{section}{0}
\renewcommand{\thefigure}{S\arabic{figure}}
\renewcommand{\thetable}{S\arabic{table}}
\renewcommand{\theequation}{S\arabic{equation}}
\setcounter{page}{1}

\section*{Supplementary Materials}
\appendix

\section{Additional Implementation Details} \label{supp:additional_implementation_details}

Our default encoder, PointBERT~\cite{Yu2022PointBERT}, is only capable of accepting fixed input sizes, which ranges from 8192 to $\sim$10k points. As the number of points in on real-world objects are much lesser, a small architectural modification is needed to accommodate them simultaneously. Point-BERT divides the point cloud into overlapping patches. To keep the in-patch (number of points and density) and between-patch (overlap amount, coverage) distribution of the divided point cloud, the number of patches are calculated individually, and a special padding token is used for the remaining patch embeddings, which are masked out in the attention mechanism of the transformer encoder part. This provides a fair baseline, as only true, meaningful points contribute to the final result. These masked patch tokens are also left out of the final global feature pooling to avoid influencing results.

\begin{table}[h!]
    \renewcommand{\arraystretch}{1.3}
    \centering
    \small
    \resizebox{\columnwidth}{!}{%
    \begin{tabular}{lcccccc}
    \toprule
    Model & Optimizer & \makecell{Peak\\LR} & \makecell{LR\\Schedule} & \makecell{Weight\\Decay} & \makecell{Batch\\Size} & Epochs \\
    \midrule
    Point-BERT & AdamW & 1e-3 & Cosine Decay & 0.1 & 512 & 250 \\
    \bottomrule
    \end{tabular}
    }
    \caption{Hyperparameters used for training.}
    \label{tab:hyp_params}
\end{table}

All models are bootstrapped from pre-trained weights rather than trained from scratch. This follows the proceedings of the state-of-the-art method ULIP-2~\cite{Xue2024ULIP2}. Point-BERT~\cite{Yu2022PointBERT} is loaded with the checkpoint obtained after its masked-reconstruction pre-training phase. 
The hyperparameters used for training are described in~\Cref{tab:hyp_params} 

\begin{table*}[h]
\centering
\renewcommand{\arraystretch}{1.2}
\small
\resizebox{1.98\columnwidth}{!}{%
\begin{tabular}{l c ccccccccccc}
\toprule
Method & Pre-train Dataset & Avg. & Car & Barrier & Ped. & Truck & T.C. & Trailer & Bus & Motor. & Bicycle & C.V. \\
\hline
\rowcolor[rgb]{0.91,0.91,0.91} Frequency (\%) & & -- & 44.9 & 18.1 & 13.7 & 8.6 & 7.0 & 2.3 & 1.7 & 1.4 & 1.3 & 1.1 \\
\hline
PointClip$^*$~\cite{zhang2022pointclip} & nuS & 21.1 & 18.8 & 2.1 & \underline{74.0} & 0.0 & 29.7 & \textbf{57.0} & 5.5 & 4.5 & 17.9 & 1.9 \\
Clip2Point$^*$~\cite{huang2023clip2point} & nuS & 24.0 & 26.7 & 10.5 & 45.2 & 16.8 & \underline{34.2} & 13.9 & 51.2 & 5.7 & 15.8 & 20.0 \\
CLIP$^{2*}$~\cite{zeng2023clip2} & nuS & 28.7 & 41.9 & \underline{17.3} & 40.3 & 41.3 & \textbf{35.3} & 20.6 & 22.5 & 22.4 & 21.1 & 24.8 \\
LidarCLIP$^{\dagger}$~\cite{hess2024lidarclip} & nuS & 10.1 & 13.7 & \textbf{20.8} & 15.7 & 5.5 & 0.1 & 0.0 & 0.5 & \underline{43.0} & 0.0 & 1.8 \\
ULIP-2~\cite{Xue2024ULIP2} & nuS & \underline{43.8} & \underline{49.7} & 15.4 & \textbf{74.9} & \underline{41.6} & 18.2 & \underline{52.0} & \underline{73.6} & 36.1 & \underline{28.0} & \textbf{49.1} \\
ULIP-2$^{\dagger}$~\cite{Xue2024ULIP2} & Objaverse~\cite{Deitke2023Objaverse} & 9.9  & 1.1 & 3.1 & 43.6 & 1.8 & 18.4 & 0.2 & 10.0 & 15.7 & 5.1 & 0.0 \\
\midrule
BlendCLIP (Ours) & Objaverse~\cite{Deitke2023Objaverse}+nuS & \textbf{47.9} & \textbf{76.2} & \underline{17.3} & 67.9 & \textbf{68.0} & 30.1 & 22.1 & \textbf{81.2} & \textbf{49.7} & \textbf{32.8} & \underline{34.2} \\
\bottomrule
\end{tabular}
}
\caption{Per-class Top-1 zero-shot classification accuracy (\%) on nuScenes. Classes are ordered by frequency of occurrence (shown in the first row). Ped: pedestrian, Motor.: motorcycle, T.C.: traffic cone, C.V.: construction vehicle. $*$: reported by authors of CLIP$^2$~\cite{zeng2023clip2}. $^{\dagger}$: reproduced by us. Best results in bold, second-best are underlined.}
\label{tab:zero_shot_classification_classwise_detailed}
\end{table*}


\section{Class-Wise Comparison:}
\label{supp_classwise_comparison}
\begin{figure}[h!] 
    \centering
    \includegraphics[width=0.99\columnwidth]{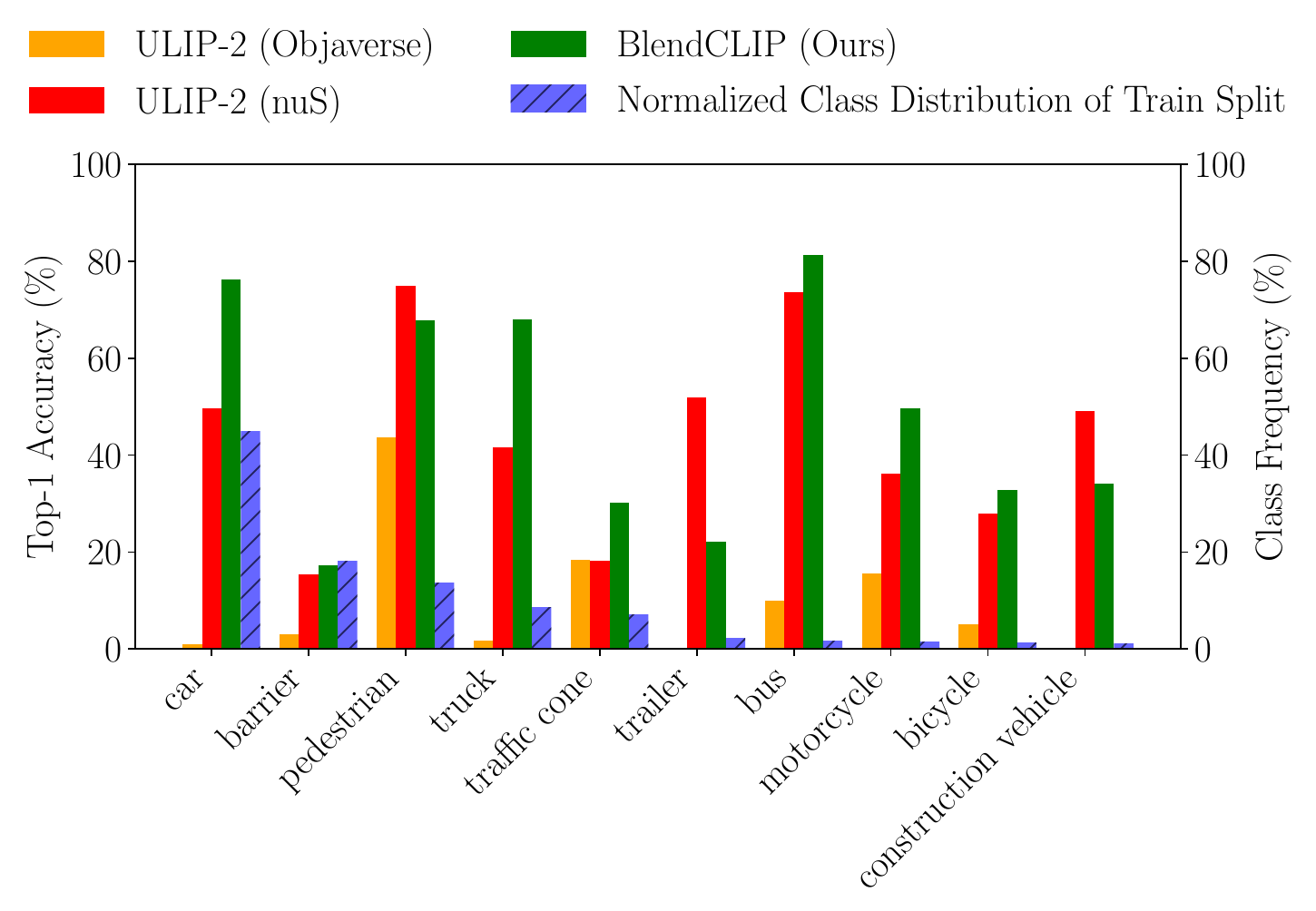} 
    \caption{Per-class top-1 zero-shot accuracy on the nuScenes-triplets validation set, with class frequencies shown in blue. Compared are ULIP-2 (Objaverse), ULIP-2 (nuScenes), and BlendCLIP (ours).} 
    \label{fig:freq_vs_acc}
    \vspace{-0.5em}
\end{figure}

To better understand how our method performs across different object categories, we analyze the class-wise top-1 accuracy on the nuScenes-triplets validation split. \Cref{fig:freq_vs_acc} compares the per-class classification accuracy with the corresponding class frequencies in the training set. As is typical for autonomous driving data, the class distribution is heavily long-tailed, with \texttt{car} making up over 40\% of all instances, while classes like \texttt{bicycle} or \texttt{construction vehicle} appear in fewer than 1\% of examples.

\suggest{Interestingly, both ULIP-2 variants exhibit only a weak dependence on class frequency, suggesting that point cloud–language alignment in general mitigates some of the long-tail effects common in supervised training. However, important differences remain. ULIP-2 trained only on Objaverse achieves good performance on semantically rich categories but collapses on outdoor-specific classes such as \texttt{construction vehicle} and \texttt{trailer}. Conversely, ULIP-2 trained on nuScenes adapts to the outdoor domain but lags behind on less frequent categories. In contrast, BlendCLIP combines the strengths of both: it maintains competitive accuracy on common classes while significantly boosting performance on rare ones (e.g., \texttt{bicycle}, \texttt{construction vehicle}). This balanced behavior indicates that our curriculum-based mixing strategy effectively transfers semantic richness from synthetic data while adapting to LiDAR-specific sensor characteristics, without requiring explicit frequency reweighting.

Table~\ref{tab:zero_shot_classification_classwise_detailed} extends this comparison to a broader set of baselines. BlendCLIP achieves the best or second-best accuracy in nearly all categories, confirming the balanced gains suggested by the histogram. Notably, it delivers large improvements on common classes such as \texttt{car} and \texttt{truck}, while also setting new best results on rarer categories like \texttt{motorcycle} and \texttt{bicycle}. Other baselines occasionally peak on individual classes (e.g., PointCLIP on \texttt{trailer}, ULIP-2 on \texttt{pedestrian}), but none match BlendCLIP’s overall consistency.
}

\suggest{\section{Additional Datasets} \label{supp:additional_datasets}
Our method is additionally evaluated on ScanObjectNN~\cite{Uy2019ScanObjectNN}, an object-centric real-world dataset collected with RGB-D sensors across 15 object categories, and ModelNet40~\cite{wu20153dShapeNets}, which is a well-established synthetic CAD object model dataset over 40 categories.

\Cref{tab:zero_shot_classification_additional} shows our method performs comparably on zero-shot classification benchmarks to current state-of-the-art methods. The performance on ScanObjectNN highlights that our method is not restricted to only outdoors LiDAR scenarios: it can be used for indoor use cases, where object categories and capture quality differ.
}

\begin{table}[h!]
    \centering
    \renewcommand{\arraystretch}{1.4}
    \small
    \resizebox{0.98\columnwidth}{!}{%
    \begin{tabular}{lccc}
        \toprule
        Method & Pre-train Dataset & ScanObjectNN~\cite{Uy2019ScanObjectNN} & ModelNet40~\cite{wu20153dShapeNets} \\
        \midrule
        PointCLIP$^*$~\cite{zhang2022pointclip} & \multirow{2}{*}{2D inference}  & 10.5 / 30.6 & 19.3 / 34.8\\
        PointCLIP v2$^*$~\cite{zhu2023pointclipv2} & & 42.2 / 74.5 & 63.6 / 85.0 \\
        CLIP2Point$^*$~\cite{huang2023clip2point} & ShapeNet & 25.5 / 59.4 & 49.5 / 81.2\\
        OpenShape-PointBERT$^*$~\cite{liu2023openshape} & Ensembled$^\dagger$ & 56.7 / \textbf{88.6} & \textbf{84.4} / \underline{98.0} \\
        ULIP-2~\cite{Xue2024ULIP2} & Objaverse~\cite{Deitke2023Objaverse} & \underline{62.3} / 85.9 & 82.4 / 97.6\\
        \midrule
        BlendCLIP (Ours) & \makecell{Objaverse~\cite{Deitke2023Objaverse}\\+ nuScenes-triplets} & \textbf{63.2} / \underline{86.1} & \underline{82.5} / \textbf{98.3}\\
        \bottomrule
    \end{tabular}
    }
    \caption{Object-wise zero-shot classification accuracies (top-1/top-5, \%) for different methods on ScanObjectNN and ModelNet40. Higher is better. Best scores are highlighted in bold, second-best are underlined. $*$: Reported by authors of OpenShape. $\dagger$: Generated by authors of OpenShape.}
    \label{tab:zero_shot_classification_additional}
\end{table}

\begin{figure}[htbp]
    \centering
    \includegraphics[width=0.8\columnwidth]{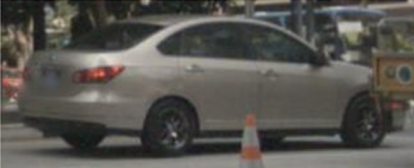}
    \caption{Example image crop used for captioning evaluation.}
    \label{fig:captiontest_car}
\end{figure}

\suggest{\section{Effect of VLM for Caption Generation}
\label{supp:caption_vlm}
For caption generation of outdoor triplets, prior works such as ULIP-2~\cite{Xue2024ULIP2} rely on BLIP-2 with OPT~\cite{zhang2022opt} as the language model head. In this experiment, we investigate replacing the VLM with the recently released Gemma 3 4B~\cite{team2025gemma}, an open-weight, instruction-tuned language model that runs efficiently on local hardware. Its explicit support for system prompts allows controlled caption style and content. Specifically, the following system prompt is used to guide Gemma’s output during caption generation:



\begin{quote}
"You are to caption images. Capture as much detail and semantic information as possible. Only describe one object, which is the largest one in the image. Ignore the background. Leave out image quality description from the caption."
\end{quote}

This prompt ensures object-centric captions, avoids visual noise from background clutter, and discourages style-based hallucinations such as “a blurry photo.”
In~\Cref{tab:caption_comparison}, we compare captions generated by BLIP-2-OPT~\cite{li2023blip2} and Gemma 3 on the same input crop, shown in~\Cref{fig:captiontest_car}. While BLIP-2 fails to correctly identify the object and offers vague descriptions, Gemma 3 provides a detailed and semantically rich caption consistent with the vehicle in the image.}

\begin{table}[h!]
    \renewcommand{\arraystretch}{1.2}
    \centering
    \small
    \resizebox{\columnwidth}{!}{%
    \begin{tabular}{l|p{0.67\columnwidth}}
    \toprule
    Model & Caption \\
    \midrule
    BLIP-2-OPT 6.7B~\cite{li2023blip2} & a photo of \underline{traffic cones} in front of a \textbf{white} car \\
    Google Gemma 3 4B~\cite{team2025gemma} & The image shows a \textbf{beige} \underline{sedan} with a \textbf{rounded} roofline, a \textbf{black} trunk, and \textbf{dark tinted} windows. It has a \textbf{chrome} strip along the bottom and \textbf{black} wheels with \textbf{silver} rims. \\
    \bottomrule
    \end{tabular}
    }
    \caption{Comparison of generated captions. \underline{Underlined} = subject; \textbf{bold} = fine-grained descriptors.}
    \label{tab:caption_comparison}
\end{table}

\suggest{\Cref{tab:effect_vlm} reports quantitative comparison between the BLIP-2 versus Gemma~3. Across all benchmarks, the differences are within $\pm0.5\%$ for LVIS and nuScenes, and within $\pm0.7\%$ for TruckScenes. This consistency indicates that the choice of captioning VLM has only marginal influence on downstream performance. We therefore attribute the majority of the observed gains to our curriculum-based data mixing strategy rather than to the specific caption generator employed.
}
\begin{table}[h]
    \centering
    \renewcommand{\arraystretch}{1.2}
    \small
    \resizebox{0.85\columnwidth}{!}{%
    \begin{tabular}{lcccc}
        \toprule
        VLM & Objaverse-LVIS~\cite{Deitke2023Objaverse} (\%) ↑ & nuS (\%) ↑ & TruckS (\%) ↑ \\
        \midrule
        BLIP-2~\cite{li2023blip2} & 46.7 / 75.2 & 48.0 / 79.6 & 25.5 / 63.5 \\
        Gemma 3 4B~\cite{team2025gemma} & 46.4 / 75.1 & 48.5 / 83.1 & 24.8 / 63.8 \\
        \bottomrule
    \end{tabular}}
    \caption{Zero-shot top-1 / top-5 classification accuracy (\%) with captions generated by BLIP-2 or Gemma 3. Performance remains stable across VLMs.}
    \label{tab:effect_vlm}
\vspace{-1em}
\end{table}


\section{Viewpoint-Aware Occlusion Synthesis} \label{supp:occlusion_synthesis}
To simulate the geometric gap between full CAD models and partial real-world LiDAR scans at training time, we apply a lightweight visibility filter that generates partial CAD views mimicking realistic occlusions. Referred to as occlusion augmentation, we hypothesize this encourages the encoder to learn from incomplete observations, as in real-world sensing. 



\begin{figure}[htbp]
  \centering
  \begin{subfigure}[b]{0.45\columnwidth}
    \centering
    \includegraphics[width=\linewidth]{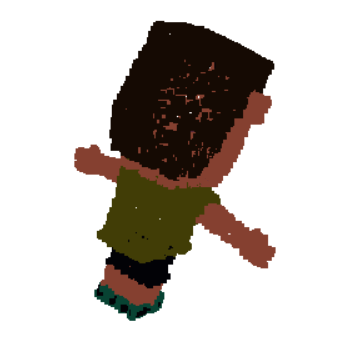}
    \caption{Before HPR}
    \label{fig:model_intact}
  \end{subfigure}
  \hfill
  \begin{subfigure}[b]{0.45\columnwidth}
    \centering
    \includegraphics[width=\linewidth]{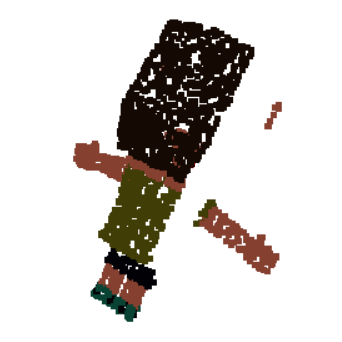}
    \caption{After HPR}
    \label{fig:model_occluded}
  \end{subfigure}
  \caption{
    Visualization of occlusion-aware augmentation using the HPR operator~\cite{katz2007direct}. The raw synthetic point cloud (a) is filtered from a sampled viewpoint to retain only the visible surface (b), simulating realistic LiDAR-style occlusion.
  }
  \label{fig:hpr_before_after}
\end{figure}

Our approach uses the Hidden Point Removal (HPR) operator~\cite{katz2007direct} to generate viewpoint-conditioned partial point clouds from synthetic objects.
HPR approximates visibility from a given virtual viewpoint without requiring expensive ray tracing. To ensure the model learns a diverse range of visibility patterns, we randomize the virtual viewpoint for each instance, sampling its position on a spherical shell around the object. This placement simulates plausible LiDAR sensor viewpoints from varying angles and distances, enhancing geometric diversity across batches. An example is shown in~\Cref{fig:hpr_before_after}.

While the HPR operator requires setting an inversion-sphere radius, we deliberately use a generously scaled radius to preserve fine-grained but partially concave features such as undercarriage cavities and wheel wells. This ensures the encoder retains access to informative geometry while still learning realistic visibility cues. The result is an occlusion-aware sampling of CAD points that attempts to better match the sparsity and perspective-induced occlusions found in real LiDAR data.

\newcommand{\cmark}{\ding{51}} 
\newcommand{\xmark}{\ding{55}}
\begin{table}[h!]
    \centering
    \renewcommand{\arraystretch}{1.4}
    \small
    \resizebox{0.85\columnwidth}{!}{%
    \begin{tabular}{cccccc}
        \toprule
        \makecell[c]{Dataset\\Mixing} & \makecell[c]{Occlusion\\Augmentation} & Objaverse-LVIS~\cite{Deitke2023Objaverse} & nuS & TruckS \\
        \midrule
        \xmark & \xmark & \textbf{47.8} & 9.9 & 8.5 \\
        \xmark & \cmark & 45.8 & 8.9 & 10.9 \\
        \cmark & \xmark & 46.4 & \textbf{48.5} & \textbf{24.8} \\
        \cmark & \cmark & 43.4 & 48.3 & 24.1 \\
        \bottomrule
    \end{tabular}
    }
    \caption{Effect of occlusion augmentation on top-1 zero-shot classification accuracy. Higher indicates better performance.}
    \label{tab:component_effectiveness}
\end{table}


\Cref{tab:component_effectiveness} compares the effectiveness of occlusion augmentation towards zero-shot classification. 
Occlusion augmentation alone provides small gains on TruckScenes but has limited effect elsewhere. When combined with mixing, performance slightly drops compared to mixing alone (–0.2\% on nuScenes, –0.7\% on TruckScenes), likely due to added geometric variability that does not fully match real-world occlusion patterns. 
These results suggest that while viewpoint-aware sparsity may help in principle, domain adaptation benefits more from capturing distributional diversity and semantic alignment than from geometric realism alone.

\suggest{\section{Qualitative Retrieval Results}
We show that our method not only performs well on general outdoor classes but can pinpoint fine-grained object attributes. There are cases where knowing the exact type of the instance is essential for correct decision-making. One such case is recognizing police officers controlling traffic or pedestrians with mobility impairments whose moving trajectories could be wildly different compared to other pedestrians.

To show this capability, we compute the similarities between the prompt and the point clouds, as described in \Cref{subsec:evaluation_protocol}, and then select the top-5 most similar examples. From the top-5 retrieved samples, we present the images of three representative instances for visualization, as images provide more intuitive interpretability than uncolored point clouds. For illustration, our qualitative analysis focuses on categories where fine-grained recognition is particularly relevant. Our choices fell to \texttt{police officer, wheelchair, stroller, motorcyclist, bicyclist}, and \texttt{scooter}. Some retrieved samples (e.g., for \texttt{stroller}) are not exact matches but semantically similar, revealing the model’s ability to surface rare, functionally relevant instances beyond the labeled taxonomy. We believe that by recognizing these rare cases, a perception system could create better educated decisions, making autonomous driving potentially safer.

In some cases, the same instance appears multiple times, as the retrieval process may independently select different point clouds of that instance. Since our method does not exploit temporal information, this recurrence demonstrates consistency: the model reliably categorizes the same instance across different times and viewpoints.}

\label{supp:retrieval}
\begin{figure*}[t]
  \centering
  \captionsetup[subfigure]{justification=centering,skip=2pt}

  \newlength{\imgheight}\setlength{\imgheight}{0.14\textheight} 
  \newlength{\imgsep}\setlength{\imgsep}{0.35em}                

  \begin{subfigure}{\textwidth}\centering
    \makebox[0pt][c]{%
      \includegraphics[height=\imgheight]{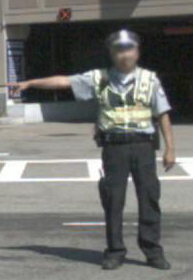}%
      \hspace{\imgsep}%
      \includegraphics[height=\imgheight]{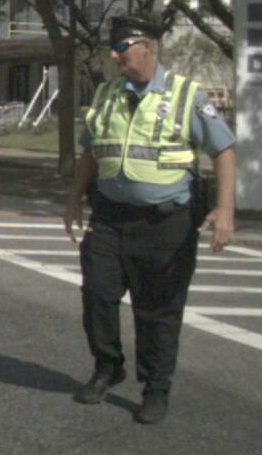}%
      \hspace{\imgsep}%
      \includegraphics[height=\imgheight]{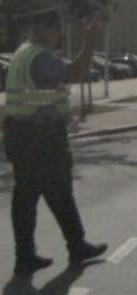}%
    }
    \subcaption{\texttt{police officer}}
  \end{subfigure}\par\vspace{0.4em}

  \begin{subfigure}{\textwidth}\centering
    \makebox[0pt][c]{%
      \includegraphics[height=\imgheight]{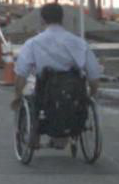}%
      \hspace{\imgsep}%
      \includegraphics[height=\imgheight]{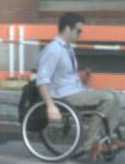}%
      \hspace{\imgsep}%
      \includegraphics[height=\imgheight]{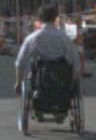}%
    }
    \subcaption{\texttt{wheelchair}}
  \end{subfigure}\par\vspace{0.4em}

  \begin{subfigure}{\textwidth}\centering
    \makebox[0pt][c]{%
      \includegraphics[height=\imgheight]{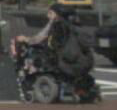}%
      \hspace{\imgsep}%
      \includegraphics[height=\imgheight]{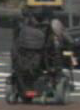}%
      \hspace{\imgsep}%
      \includegraphics[height=\imgheight]{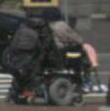}%
    }
    \subcaption{\texttt{stroller}}
  \end{subfigure}\par\vspace{0.4em}

  \begin{subfigure}{\textwidth}\centering
    \makebox[0pt][c]{%
      \includegraphics[height=\imgheight]{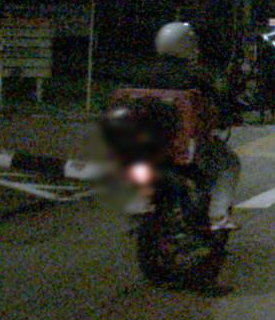}%
      \hspace{\imgsep}%
      \includegraphics[height=\imgheight]{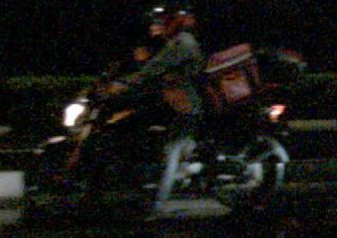}%
      \hspace{\imgsep}%
      \includegraphics[height=\imgheight]{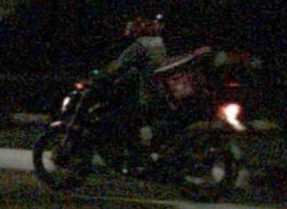}%
    }
    \subcaption{\texttt{motorcyclist}}
  \end{subfigure}\par\vspace{0.4em}

  \begin{subfigure}{\textwidth}\centering
    \makebox[0pt][c]{%
      \includegraphics[height=\imgheight]{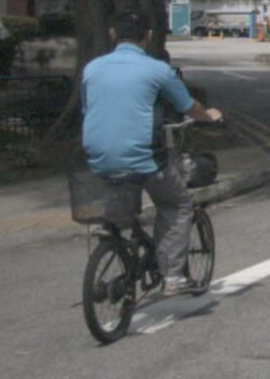}%
      \hspace{\imgsep}%
      \includegraphics[height=\imgheight]{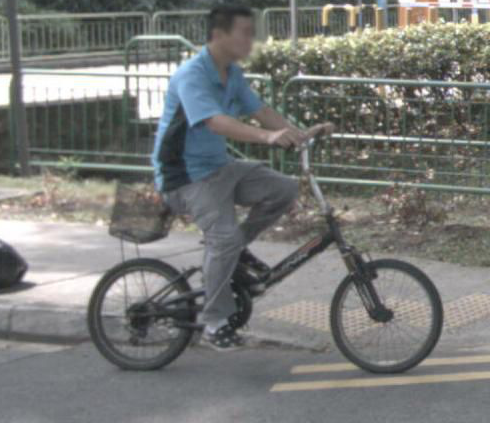}%
      \hspace{\imgsep}%
      \includegraphics[height=\imgheight]{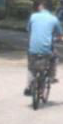}%
    }
    \subcaption{\texttt{bicyclist}}
  \end{subfigure}\par\vspace{0.4em}

  \begin{subfigure}{\textwidth}\centering
    \makebox[0pt][c]{%
      \includegraphics[height=\imgheight]{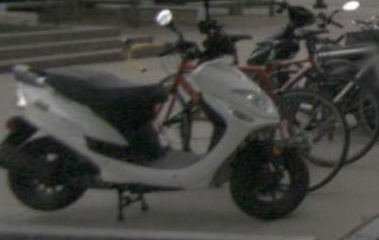}%
      \hspace{\imgsep}%
      \includegraphics[height=\imgheight]{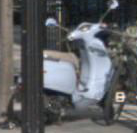}%
      \hspace{\imgsep}%
      \includegraphics[height=\imgheight]{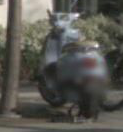}%
    }
    \subcaption{\texttt{scooter}}
  \end{subfigure}
    
    \caption{Selected examples from top-5 retrieved samples for BlendCLIP. The images are shown only for visualization. The similarities were computed only with the embeddings of the point clouds. Every image has a different point cloud associated with it.}
  \label{fig:retrieval}
\end{figure*}

\end{document}